# Fake Hilsa Fish Detection Using Machine Vision

Mirajul Islam[1], Jannatul Ferdous Ani[1], Abdur Rahman[1], Zakia Zaman[1]

[1] Department of Computer Science and Engineering, Daffodil International University, Dhaka 1027, Bangladesh
{merajul15-9627, jannatul15-10483, abdur15-10022, zakia.cse}@diu.edu.bd

**Abstract.** Hilsa is the national fish of Bangladesh. Bangladesh is earning a lot of foreign currency by exporting this fish. Unfortunately, in recent days some unscrupulous businessmen are selling fake Hilsa fishes to gain profit. The Sardines and Sardinella are the most sold in the market as Hilsa. The government agency of Bangladesh, namely Bangladesh Food Safety Authority said that these fake Hilsa fish contain high levels of cadmium and lead which are detrimental for humans. In this research, We have proposed a method that can readily identify original Hilsa fish and fake Hilsa fish. Based on the research available on online literature, we are the first to do research on identifying original Hilsa fish. We have collected more than 16,000 images of original and counterfeit Hilsa fish. To classify these images, we have used several deep learning based models. Then the performance has been compared between them. Among those models, DenseNet201 achieved the highest accuracy of 97.02%.

## 1 Introduction

Hilsa fish is known as one of the tastiest and favorite fish in the world. It is named as Ilish in Bangladesh. Every year Bangladesh produces 75% of the total Hilsa fish production in the world [1]. And the other 25% produced in India, Myanmar, Pakistan, and some countries along the Arabian sea. Three kinds of Hilsa fishes are available in Bangladesh. These are Tenualosa Ilisha which is called Padma Ilish and it is the most popular Hilsa fish in the world. Another is Nenuacosa Toli, which is called Chandana Ilish. And the last one is Hilsha Kelle that is known as Gurta Ilish. Hilsa is known as saltwater fish, but it lays eggs in freshwater rivers (Padma, Meghna, Jamuna) delta at the Bay of Bengal. The demand for Hilsa fish in Asian countries exists throughout the year. However, it increases a lot especially in festivals like Pohela Boishakh (Bengali new year), Saraswati Puja, etc.

In many countries, the production of Hilsa fish is now declining. But its production in Bangladesh is increasing every year. By exporting these fishes in many countries of the world, a large amount of foreign currency is earned every year. It adds 1% of the total GDP in Bangladesh [1]. For several years in many countries, including

Bangladesh, India, some unscrupulous traders are selling fake Hilsa fish. With a large portion of original Hilsa fish, they added part of fake Hilsa fish inside it then exported to many countries. This tarnishes the image of our country and in return lowering the revenues.

These fake Hilsa look a lot like the original one, but there are some differences in shape, mostly in the head and tail, and also in the taste and smell. The Sardine, Sardinella, Cokkash, Chapila, Indian oil sardine, all of these look like original Hilsa fish. The body of the Hilsa fish is equally long towards the abdomen and back but the abdomen is higher than the back in Sardine. The Head of the Hilsa fish is slightly longer, on the other hand, the head of the sardine fish is slightly smaller and the front side is slightly blunt. Fake Hilsa's eyes are bigger than the real Hilsa fish. The back of the original Hilsa is bluish green. Sardine has black dots at the base of the dorsal fin, but Hilsa fish has black dots on its gill. The front of the sardine's back fins and the edges of the tail fins are blurred but the hind fins of Hilsa are whitey.

But the sad truth is, fake Hilsa can not be easily detected by looking at it many times. To solve this problem we use machine vision based deep learning models to identify the original fish. Our proposed method can easily classify Hilsa fish and fake Hilsa fish with high accuracy. We used Xception, VGG16, InceptionV3, NASNetMobile, and DenseNet201 to find out which model gives better performance.

The remainder of this article is organized as follows: Section 2 describes some related work. Section 3 describes the research methodology and a very brief study on the used models. Section 4 shows the result analysis. And section 5 concludes the study with future direction.

## 2  Background Study

We studied some notable research work on fish classification and also on image processing. Pavla et al. [2] used a modified Rosenblatt algorithm to classify six species of fish. They have used 2132 silhouettes images. Of these, 1406 images were used for model training, and 726 used for model testing. Out of 726 images, their modified model was able to classify 386 images correctly. Israt et al. [3] used SVM, K-NN, Ensemble-based algorithm to recognize six types of local fish. They used a histogram based method to segment the gray-scale fish images. In their research, SVM achieved the highest accuracy of 94.2%. In the paper [4], the authors used pre-trained VGG16 for feature extraction of eight species of fish images and logistic regression has been used for the classification. Then they achieved 93.8% accuracy. Rowell et al. [5] used the InceptionV3 model to classify Nile tilapia is harvested alive or Hibay. They used Adam optimizer in the last two layers of the model for increasing the accuracy rate. After 1000 iterations their model achieved the best accuracy. Hafiz et al. [6] used 915 fish images from six different classes. Six different types of CNN model has been used in this research namely, VGG-16 for transfer learning, one-block VGG, two-block VGG, three-block VGG, LeNet 5, AlexNet, GoogleNet, ResNet 50, and a proposed 32 Layer CNN with 404.4 million parameters, which gives the best accuracy than the rest. Xiang et al. [7] used transferred DenseNet201 for the diagnosis of breast abnormality.

Here, 114 abnormal tissue images have been used for classification. Then their model gave 92.73% accuracy. Jing et al. [8] used a multiclass support vector machine(MSVM) algorithm to classify six species of freshwater fish. They have captured all the images in 1024 X 768 size. Then multiple features have been extracted from these images. Ogunlan et al. [9] used a Support Vector Machine (SVM)-based technique for classifying 150 fish. Of these, 76 have been used for model training and 74 for testing the model. They have used 3 more models in their research. Those are Artificial Neural Networks (ANNs), K-means Clustering, and K-Nearest Neighbors (KNN). And Principal Component Analysis (PCA) is used to reduce the dimensionality of a dataset. SVM achieved the highest accuracy of 74.32%. Mohamad et al. [10] used the SVM, KNN, and ANN models to classify Nile tilapia fish. They used the Scale Invariant Feature Transform (SIFT) and Speeded Up Robust Features (SURF) algorithms for feature extraction. SURF achieved the highest accuracy of 94.44%. However, no prior work has been done on identifying fake Hilsa fish and the image dataset used in this study is also not available before our research.

## 3 Research Methodology

The following Fig. 1 shows the working strides of our study. In brief, after assembling the datasets, it is labeled, pre-processed, and divided into the train and validation set. Using the datasets, five different CNN models are trained and validated to evaluate model performance. Their performance has been visualized through the confusion matrix and several evaluation metrics such as accuracy, precision, recall, F1 score are used to compare among different models.

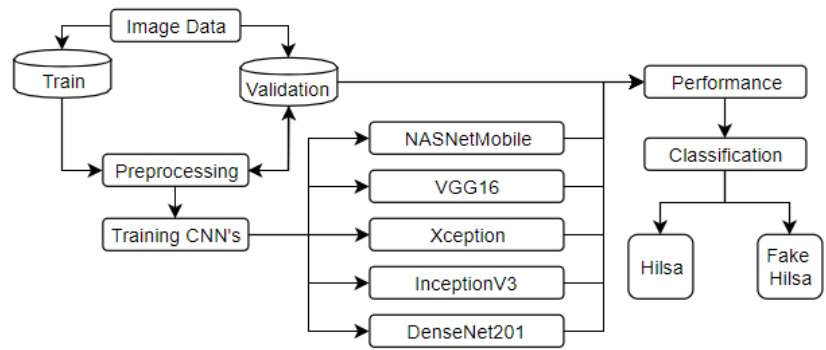

**Fig. 1.** Proposed classification procedure.

### 3.1 Image Data Collection and Pre-process

There have been 16,622 different sizes of images that have been categorized into two groups named real Hilsa and fake Hilsa. Those different sizes of images are taken from the endemic fish market and many alternative sources. There are three different types of Hilsa fish in the 8,722 images: Tenualosa Ilisha (Padma Ilish), Nenuacosa Toli (Chandana Ilish), and Hilsha Kelle (Gurta Ilish). The other 7,900 images are of different types of Fake Hilsa fish. Mostly Sardine, Sardinella, Indian oil sardine, Cokkash, Chapila, and few other fishes that look similar to Hilsa. All 16,622 images were resized into 224 x 224 x 3 to train the model. All the collected images are divided into training and validation data set. 80% (13,301) images are in the training set and another 20% (3,321) images are in the validation (Fig. 3). We have named the test set as a validation set. And labeled our dataset into 2 classes, namely Class 0 (Fake Hilsa Fish) and Class 1 (Hilsa Fish) shown in Fig. 2.

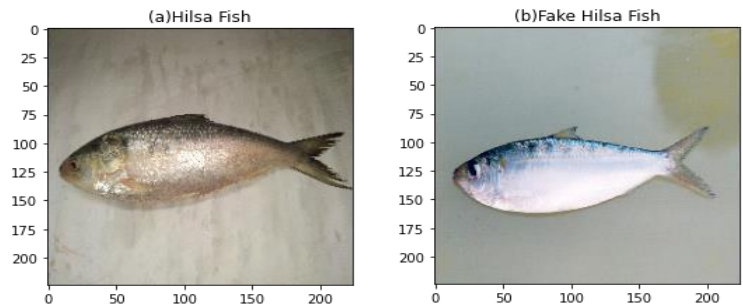

**Fig. 2.** Images of 2 classes (a) Hilsa Fish (b) Fake Hilsa Fish

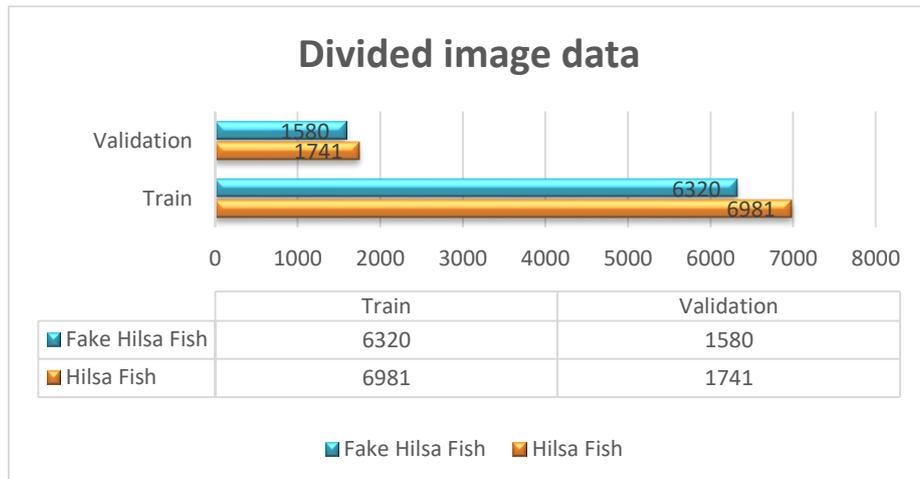

**Fig. 3.** The image dataset is divided into training and validation sets.

### 3.2 Model Generation

We use five different types of CNN models to identify the original Hilsa fish and the fake one. The characteristics of these CNN models [11] are shown in Table 1. A brief discussion of these models is given below.

Convolution Neural Network (CNN) is a type of Deep Neural Network (DNN) which applies for image processing, natural language processing, and many other classifications, recognition, and detection tasks. It includes an input layer, some hidden layers known as convolution layers, pooling layers, flattening layers, and fully connected layers. The convolution layer is the most significant layer for extracting features, edges, colors, shape, pattern, etc from the images. The output of this layer is the feature map. Pooling layers reduces the size of an image. Mainly three types of pooling layers are used: max, average, and sum. It removes 3/4% of the activation, seen in the previous layers. Fully connected layers (FC) are called Dense layers [12]. Fully connected layers mean all nodes in one layer connected to the output of the next layers.

**Table 1.** Model characteristics in details.

| Model | Size(MB) | Parameters | Depth |
|---|---|---|---|
| Xception | 88 | 22,910,480 | 126 |
| VGG16 | 528 | 138,357,544 | 23 |
| InceptionV3 | 92 | 23,851,784 | 159 |
| DenseNet201 | 80 | 20,242,984 | 201 |
| NASNetMobile | 23 | 5,326,716 | - |

Xception [13] was invented by google researchers. It is an architecture based on depthwise separable convolution layers. It has 36 convolution layers into 14 modules. Parameters are quite similar to Inception-V3. It has been trained with millions of images from the Imagenet database [14].

VGG16 [15] is sometimes called Oxfordnet because it was invented by a visual geometry group from Oxford in 2014. It was first used to win the ILSVRC (Imagenet [14]) competition. The input shape is fixed for the conv1 layer (224 x 224 x 3). Although it's size is quite larger than the rest, it is very useful for learning purposes and easy to implement.

InceptionV3 [16] [17] was invented in 2015 by Google Inc. It has a total of 48 deep layers. It has been trained with millions of images from the Imagenet database [14]. InceptionV3 is the most used convolution neural network model for image recognition.

NASNetMobile [18] is another type of convolution neural network(CNN) that is divided into two cells: a normal cell that returns a feature map in the same aspect, and a reduction cell which reduces the height and width of a feature map. It has been trained with millions of images from the Imagenet database [14]. It has 3 input channels, height, width, and RGB color channel. It requires 224 x 224 x 3 input shape for images.

DenseNet201 [19] has a total of 201 deep layers. We achieved the highest accuracy by using this model. The input shape is 224 x 224 x 3 for "channel-last" data format and 3 x 224 x 224 is for "channel-first" format. The summary of DenseNet201 CNN model is too large which cannot be fully disclosed in this article. That's why we have visualized here some beginning layers (Fig. 4a) and some ending layers (Fig. 4b).

```
Model: "model_1"
Layer (type)                   Output Shape         Param #     Connected to
================================================================================
input_1 (InputLayer)           (None, 224, 224, 3)  0
zero_padding2d_1 (ZeroPadding2D (None, 230, 230, 3) 0           input_1[0][0]
conv1/conv (Conv2D)            (None, 112, 112, 64) 9408        zero_padding2d_1[0][0]
conv1/bn (BatchNormalization)  (None, 112, 112, 64) 256         conv1/conv[0][0]
conv1/relu (Activation)        (None, 112, 112, 64) 0           conv1/bn[0][0]
zero_padding2d_2 (ZeroPadding2D (None, 114, 114, 64) 0          conv1/relu[0][0]
pool1 (MaxPooling2D)           (None, 56, 56, 64)   0           zero_padding2d_2[0][0]
conv2_block1_0_bn (BatchNormali (None, 56, 56, 64)  256         pool1[0][0]
conv2_block1_0_relu (Activation (None, 56, 56, 64)  0           conv2_block1_0_bn[0][0]
```

(a) Beginning Layer.

```
batch_normalization_1 (BatchNor (None, 1024)        4096        dense_1[0][0]
activation_1 (Activation)      (None, 1024)         0           batch_normalization_1[0][0]
dropout_1 (Dropout)            (None, 1024)         0           activation_1[0][0]
dense_2 (Dense)                (None, 1024)         1049600     dropout_1[0][0]
batch_normalization_2 (BatchNor (None, 1024)        4096        dense_2[0][0]
activation_2 (Activation)      (None, 1024)         0           batch_normalization_2[0][0]
dropout_2 (Dropout)            (None, 1024)         0           activation_2[0][0]
dense_3 (Dense)                (None, 2)            2050        dropout_2[0][0]
================================================================================
Total params: 21,348,930
Trainable params: 21,115,778
Non-trainable params: 233,152
```

(b) End Layer.

**Fig. 4.** Model summary of DenseNet201.

To train those models we require to resize the images as 224 x 224 x 3 and also rescale them into 1/255 pixel values. Because customarily the pristine pixel values of the images are integers with RGB coefficients between 0-255. This slows down the learning process because the range of integer values is too large. For this, we need to normalize our pixel value between 0-1. This method is known as min-max normalization [20]. All the models are trained and the results are visualized using Scikit-learn, Keras, OpenCV, Matplotlib, and TensorFlow as backends. Google colab [21] used to execute all the processes and experiments.

### 3.3 Features Extraction

There are various convolution layers in each CNN model. Convolution layers that process the two-dimensional images are known as conv2d layers. Each conv2d gets input images, having three color channels. Then this input image is processed through a convolution filter. This filter is known as a convolution kernel(extract features of the input image) or a feature detector. Each input image is arranged in a matrix which consists of combining the value of three color channels. The matrix tends to get smaller for each filter to read the features of the input image, therefore there are many differences to see the same image in each filter. High-level filters read the special pattern of input images and low-level filters read the normal features of the input image [22]. Each filter normally carries pre-defined weights. The term feature extraction means the reduction of features and RGB from the raw input image in every conv2d layer. Each conv2d has some arguments [23] or parameters for image processing, those are filter, kernel-size, strides, padding, data format, dilation-rate, groups, activation, use bias, and a few more. These arguments are different in each CNN model which we have used in our research to identify Hilsa fishes.

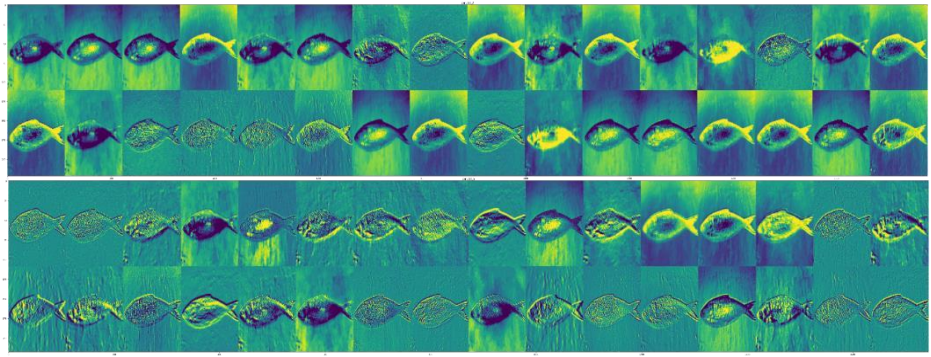

**Fig. 5.** Visualize the features extraction from the first two conv2d (DenseNet201).

In our research, 32 feature maps are extracted by each conv2d layer. We can see from Fig. 5, many changed images of the Hilsa fish after applying filters in each conv2d. Among these filters some have highlighted the edges, some have highlighted the shape, and some have highlighted the background and heat-map. CNN reads the images in each conv2d layer and learns the features of every input image for doing better classification.

## 4    Result Analysis

This section discusses the experimental results of the five CNN models. We train those models with 6981 original Hilsa and 6320 fake Hilsa fish images. Models are tested with 1741 Hilsa fish and 1580 fake Hilsa images to evaluate performance. After that NASNetMobile gave the lowest accuracy of 86.75% and DenseNet201 gave the highest accuracy of 97.02%.

As shown in Fig. 6, shows the training accuracy and loss as well as the validation accuracy and loss in each epoch of these five models. In Fig. 6a, training accuracy in the last epoch 89.1% on the other side last epoch validation accuracy 78.02% which is less than the first epoch accuracy 96.98%. The training loss at the last epoch is the lowest at 28.12% and the validation loss is 54.56%. The obtained result of this graph is for the NASNetMobile model. In Fig. 6b, train accuracy after all epoch is 94.25%, and validation accuracy is 91.19%. At the same time training and validation loss in the last epoch respectively 32.79% and 15.20%. That result is for VGG16. In Fig. 6c, 93.95% training accuracy, 96.77% validation accuracy got from the last epoch and 18.24% training loss, 13.82% validation loss got from the last epoch for the Xception model. In Fig. 6d, got 93.88% training and 95.01% validation accuracy in the last epoch. Also got 40.42% training and 32.21% validation loss for the InceptionV3 model. From Fig. 6e, 89.62% of training accuracy and 95.36% validation accuracy got from the last epoch. With this 46.16% training and 26.36% validation loss is obtained from the last epoch. That results is for the DenseNet201 model. Considering the average results of these five graphs, Fig. 6e is better than the others.

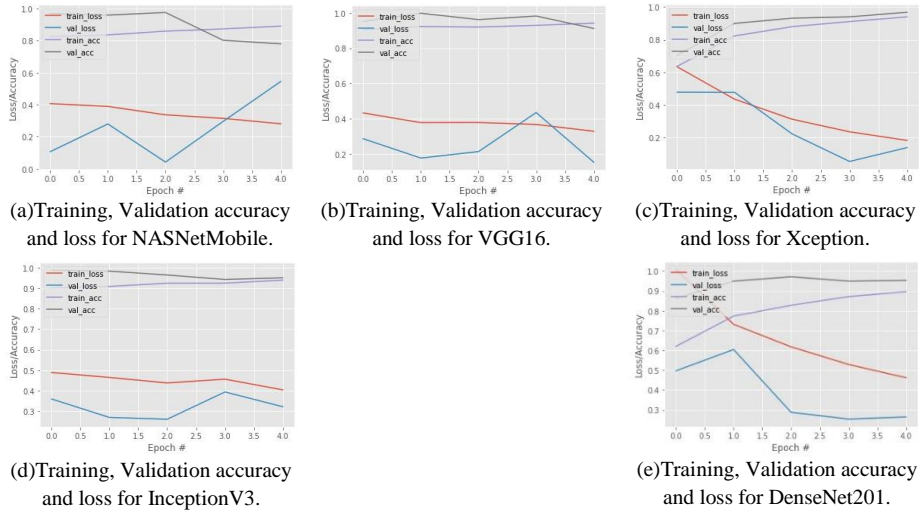

(a) Training, Validation accuracy and loss for NASNetMobile.

(b) Training, Validation accuracy and loss for VGG16.

(c) Training, Validation accuracy and loss for Xception.

(d) Training, Validation accuracy and loss for InceptionV3.

(e) Training, Validation accuracy and loss for DenseNet201.

**Fig. 6.** Classification performance.

Table 2 lists the results of the confusion matrix for the five CNN models. A confusion matrix is the measure and visualized the total right and wrong predictions made by the classifiers. When the model predicts it is a Hilsa fish and the actual output is also Hilsa fish then it is a true positive (TP). When the model predicts it is a fake Hilsa fish and the actual output is also fake Hilsa fish then that term is true negative (TN). When the classifier predicts it is a Hilsa fish but the actual output is fake Hilsa fish then it is a false positive (FP). And when the classifier predicts it as a fake Hilsa fish and the actual output is also a Hilsa fish then it is a false negative (FN).

According to Table 3, we found in NasnetMobile that the difference between the false-positive rate and the false-negative rate is high. DenseNet201 shows a significant improvement because the difference between FP and TP rate is low rather than the NasnetMobile. The accuracy measure of Xception, Inceptionv3, and DenseNet201 are very close to each other. The accuracy of Xception is 96.60% and InceptionV3 is 96.87%. Here the accuracy measured difference between them is very low which is 0.27%. The accuracy of DenseNet201 is 97.02%, the difference of its accuracy with InceptionV3 is only 0.15%. Based on the data in Table 2, it can be stated that the DenseNet201 is giving better classification performance than the other model.

Table 2. Confusion matrix result for five CNN models.

| Model | True Positive(TP) | True Negative(TN) | False Positive(FP) | False Negative(FN) | Accuracy (%) |
|---|---|---|---|---|---|
| NASNetMobile | 1331 | 1550 | 30 | 410 | 86.75 |
| VGG16 | 1568 | 1578 | 6 | 173 | 94.61 |
| Xception | 1666 | 1542 | 38 | 75 | 96.60 |
| InceptionV3 | 1652 | 1565 | 15 | 89 | 96.87 |
| DenseNet201 | 1659 | 1563 | 17 | 82 | 97.02 |

Table 3. Classification report for five CNN models.

| Model | Accuracy (%) | Precision (%) | F1 (%) | Sensitivity (%) | Specificity (%) | FPR (%) | FNR (%) |
|---|---|---|---|---|---|---|---|
| NASNetMobile | 86.75 | 97.80 | 85,82 | 79.08 | 97.79 | 1.89 | 23.54 |
| VGG16 | 94.61 | 99.62 | 94.60 | 90.09 | 99.61 | 0.37 | 9.93 |
| Xception | 96.60 | 97.77 | 96.72 | 95.36 | 97.76 | 2.40 | 4.30 |
| InceptionV3 | 96.87 | 99.10 | 96.95 | 94.61 | 99.10 | 0.94 | 5.11 |
| DenseNet201 | 97.02 | 98.99 | 97.10 | 95.01 | 98.98 | 1.07 | 4.79 |

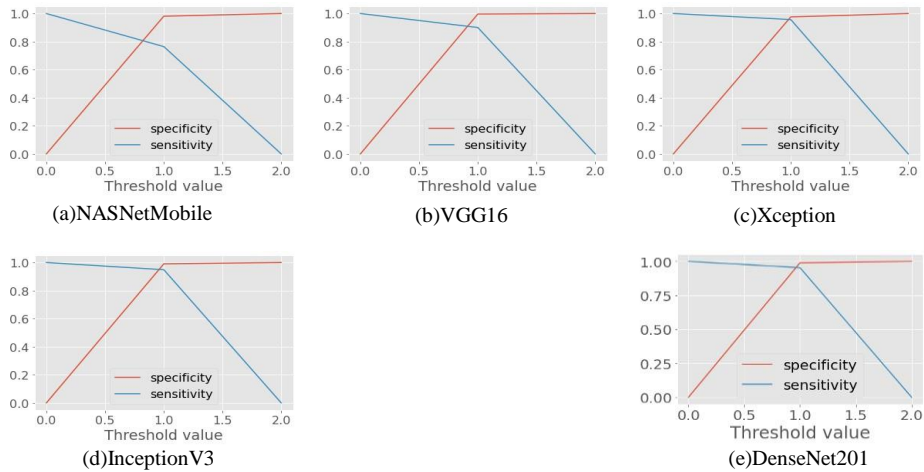

(a)NASNetMobile  (b)VGG16  (c)Xception

(d)InceptionV3  (e)DenseNet201

**Fig. 7.** Visualized sensitivity and specificity for five CNN models. (a)NASNetMobile, (b)VGG16, (c)Xception , (d)InceptionV3, (e)DenseNet201.

From Table 3, we observe that the F1 score is increasing along with the accuracy from top to bottom. Here one noticeable thing is that NASNetMobile has a difference between sensitivity and specificity, which is also visible in Fig. 7a. Apparently for Xception (Fig. 7c), InceptionV3 (Fig. 7d), and DenseNet201 (Fig. 7e) the graphs seem to look the same but we can see from Table 3, that there are some differences between their sensitivity and specificity.

In Fig. 8, the Receiver Operating Characteristics (ROC) Curve and Area Under the Curve (AUC) are visualized. ROC is a probability curve and AUC represents the degree or measure of separability between classes. When the ROC is higher that means the model is performing well. The ROC curve is plotted with TPR against the FPR where TPR is on the y-axis and FPR is on the x-axis [24]. The ROC/AUC for Xception (Fig. 8c), InceptionV3 (Fig. 8d) and DenseNet201 (Fig. 8e) is 0.97.

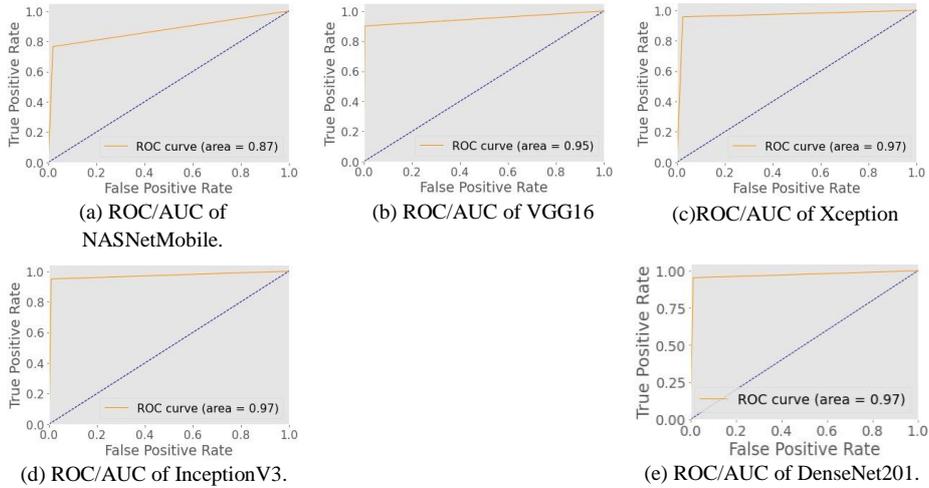

(a) ROC/AUC of NASNetMobile.  (b) ROC/AUC of VGG16  (c) ROC/AUC of Xception

(d) ROC/AUC of InceptionV3.  (e) ROC/AUC of DenseNet201.

**Fig. 8.** Receiver Operating Characteristics (ROC) Curve-Area under the curve(AUC) for five CNN models.

From the intensive result analysis, we can conclude that DenseNet201 gives a better classification performance in our experiment.

## 5  Conclusion

The main objective of this research describes the proposed methodology to identify original Hilsa fishes. This is the first study of original Hilsa and fake Hilsa identification using Convolutional neural network (CNN). We used five CNN models in our experiments and observed their performance. Then We have also performed an observational comparative analysis of our obtained results. Here, NASNetMobile shows the least performance with an accuracy of 86.75%, and DenseNet201 produces a very good performance with an accuracy of 97.02%. We hope that this study will be beneficial for researchers who will research-related topics. In the future, we want to enrich our dataset with more images and develop a mobile application that can be used to identify the original Hilsa fish in real time.